%% file: tracking-paper.tex
\documentclass[final]{cvpr}
\input{0-preamble}
\begin{document}

\title{Probabilistic Tracking with Deep Factors}

\author{Fan Jiang \hspace{0.3cm} Andrew Marmon \\
 Ildebrando De Courten \hspace{0.3cm} Frank Dellaert\\
Georgia Institute of Technology\\
Atlanta, GA\\
{\tt\small \{fan.jiang|amarmon3|icourten3|fd27\}@gatech.edu}
\and
Marc Rasi\\
\hspace{1cm} \\
Google Brain\\
Mountain View, CA\\
{\tt\small marcrasi@google.com}
}

\maketitle

\input{0-abstract}
\input{0-macros}
\input{1-intro}
\input{2-related}
\input{3-approach}
\input{4-encodings}

\input{5-results}

\input{6-conclusions}

{\small
\bibliographystyle{ieee_fullname}
\bibliography{tracking-refs}
}

\end{document}

%% file: 0-preamble.tex
\usepackage{times}
\usepackage{epsfig}
\usepackage{graphicx}
\usepackage{amsmath}
\usepackage{amssymb}

\usepackage{float}
\usepackage{xspace}
\usepackage{algorithm}
\usepackage[noend]{algpseudocode}

\usepackage[pagebackref=true,breaklinks=true,colorlinks,bookmarks=false]{hyperref}

\def\cvprPaperID{2905} 
\def\confYear{CVPR 2021}

%% file: 0-abstract.tex
\begin{abstract}
In many applications of computer vision it is important to accurately estimate the trajectory of an object over time by fusing data from a number of sources, of which 2D and 3D imagery is only one. 
In this paper, we show how to use a deep feature encoding in conjunction with generative densities over the features in a factor-graph based, probabilistic tracking framework.
We present a likelihood model that combines a learned feature encoder with generative densities over them, both trained in a supervised manner. We also experiment with directly inferring probability through the use of image classification models that feed into the likelihood formulation.
These models are used to implement deep factors that are added to the factor graph to complement other factors that represent domain-specific knowledge such as motion models and/or other prior information. Factors are then optimized together in a non-linear least-squares tracking framework that takes the form of an Extended Kalman Smoother with a Gaussian prior.
A key feature of our likelihood model is that it leverages the Lie group properties of the tracked target's pose to apply the feature encoding on an image patch, extracted through a differentiable warp function inspired by spatial transformer networks.
To illustrate the proposed approach we evaluate it on a challenging social insect behavior dataset, and show that using deep features does outperform these earlier linear appearance models used in this setting.
\end{abstract}

%% file: 0-macros.tex
\newcommand{\traj}{Z}
\newcommand{\define}{\doteq}
\newcommand{\state}{z}
\newcommand{\pose}{g}
\newcommand{\poses}{G}
\newcommand{\app}{a}
\newcommand{\apps}{A}
\newcommand{\factor}{f}
\newcommand{\unary}{\phi}
\newcommand{\binary}{\psi}
\newcommand{\RR}{R}
\newcommand{\meas}{x}
\newcommand{\feat}{c}
\newcommand{\encoder}{C}
\newcommand{\region}{R}
\newcommand{\Normal}{\mathcal{N}}
\newcommand{\mat}[1]{\ensuremath{\boldsymbol{#1}}}
\newcommand{\argmin}{argmin}

\newcommand{\fan}[1]{{\color{red}\textbf{Fan: }~#1}}
\newcommand{\frank}[1]{{\color{red}\textbf{Frank: }~#1}}
\newcommand{\andrew}[1]{{\color{red}\textbf{Andrew: }~#1}}

%% file: 1-intro.tex
\section{Introduction}

\begin{figure}[t]
    \begin{center}
       \includegraphics[width=0.9\linewidth]{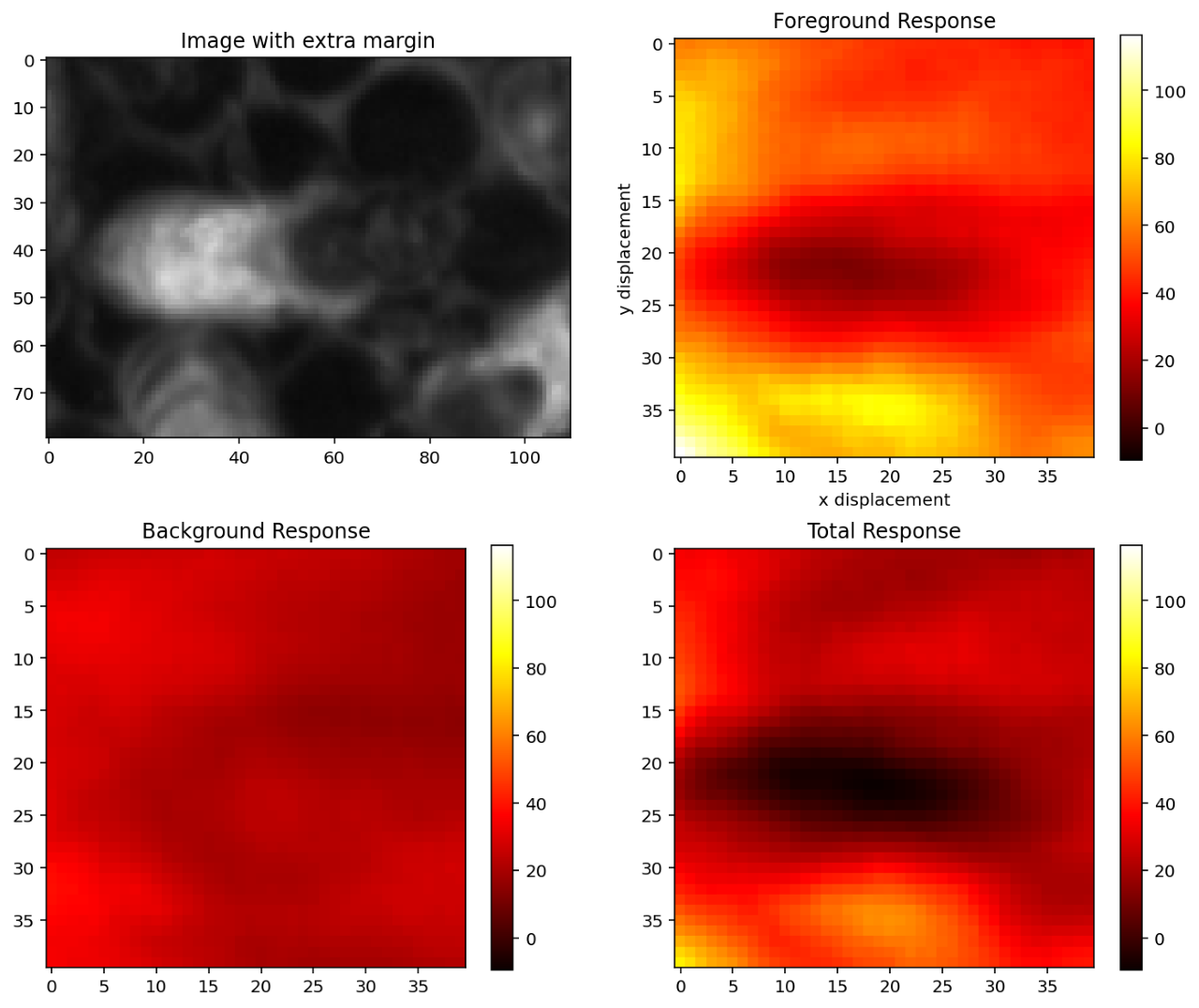}
    \end{center}
    \caption{Our proposed deep factor log-likelihood is illustrated here for a bee-tracking example. In this case, the decode stage of an auto-encoder induces a negative log-likelihood $\unary_F(\delta_x, \delta_y)$ and $\unary_B(\delta_x, \delta_y)$ for respectively the foreground and background class, here shown just for translation, although the pose $\pose$ is an element of $SE(2)$. The total response is $\unary_F(\delta_x, \delta_y) - \unary_B(\delta_x, \delta_y)$.\label{fig:idea}\vspace{-1em}}
\end{figure}

In many applications of computer vision it is important to accurately estimate the trajectory of an object over time by fusing data from a number of sources, of which imagery is only one. 
For example, when tracking a person from an autonomous vehicle, the known ego-motion of the vehicle and probabilistic motion models of the person can be brought to bear~\cite{Jung10ijsr_motion_tracking}. 
This holds as well for tracking other vehicles, for which we might have even more accurate motion information. 
Finally, and this is the application we will focus on in this paper, when tracking animals or social insects, behavior models might be available that constrain and help recover the actual movements. 

Thanks to recent advances in tracking that make use of deep learning, the usability of tracking in real world scenarios has been greatly improved. Two representative examples are Transformer Tracking~\cite{Chen21cvpr_transformer_tracking} and RPT~\cite{ma2020rpt}, both showing impressive performance.

However, modern tracking methods usually do not deal with other sources of information and rarely learn or use a motion model~\cite{Marvasti21its_survey}. Some recent work~\cite{Chen18eccv,Zhong19tip} implicitly model object motion implicitly in the form of a recurrent network, which helps ambiguity resolution in the presence of multiple similar objects. In particular, Zhong \etal~\cite{Zhong19tip} introduce the idea of iterative updating in a coarse-to-fine framework. 

In the field of animal tracking, motion models have also been used to assist in dense entity resolution for tracking~\cite{Bozek2017OIST} or to define an Region-Of-Interest (ROI) as input to infer an animal's pose~\cite{Mathis2018DeepLabCutMP}. These methods however cannot be used to inject probabilistic information about the target, and can certainly not correct for it after \textit{ex post}. Meanwhile, energy-based approaches like ours can be easily combined with other sources of information (See also Fig. 8 in~\cite{LeCun06atutorial}).

In this paper we show how to use deep feature encoding in conjunction with generative models over the features in a probabilistic tracking framework. We use factor graphs, graphical models that have become popular in state estimation~\cite{Hartley18iros_state_estimation}, navigation~\cite{Dong16rss_motion_planning}, and mapping~\cite{Teixeira16iros_underwater} to model the different sources of measurement information and probabilistic prior knowledge. 
Most factors are designed by hand, however, and only a handful of papers have leveraged data to learn the functions associated with them, see e.g.~\cite{Czarnowski20ral_deep_factors, Sodhi21corl_LEO}. 

A key feature of our likelihood model is that it leverages the Lie group properties of the tracked target's pose to apply the feature encoding on an image patch, extracted through a differentiable warp function inspired by spatial transformer networks (STNs)~\cite{Jaderberg15nips_stn}. This is similar, but different from the decoupled approach in~\cite{Clark17aaai}. We frame this is in a general way such that the pose can be any smooth manifold, including but not limited to 2D and 3D rigid transforms or similarity transforms.

While the derivative of a neural network (NN) mapping is usually only used in the incremental update of the NN parameters, there exists literature that successfully exploits the backward pass, most notably, to generating adversarial samples~\cite{Goodfellow15iclr} and incremental updating of motion estimates~\cite{Carreira16cvpr}. In our approach, we use the derivative of the neural network to implicitly carry out inference, in the the spirit of Structured Prediction Energy Networks (SPENs)~\cite{Belanger16pmlr}.

To illustrate the proposed approach we evaluate it on a challenging social insect application, a domain where previous approaches have already applied generative models~\cite{Khan04cvpr}. We show using that deep features does outperform these earlier linear appearance models, and show that simple Gaussian generative foreground and background densities suffice to obtain excellent results even when compared with modern complex networks.

%% file: 2-related.tex
\section{Related Work}

\noindent\textbf{Classical Methods:}
In recent years, the majority of state of the art object trackers have been based on two dominant methodologies: The discriminative correlation filter~\cite{Bolme10,danelljan,joao,dimp} and the Siamese correlation network~\cite{bertinetto2016fullyconvolutional,Wang19cvpr_siammask,tao2016siamese,zhang2020ocean}. 
Both methods compute a similarity metric between a target object's appearance and proposed image patches over the current frame, albeit with different architectures. 
A good review on these types of tracking methods can be found in~\cite{Fan21thesis}.
Our method similarly computes a value for similarity on the proposed image patches and additionally optimises on the best proposed patch. In particular, our method leverages the Lie-Group properties of the target's pose to compute the similarity metric and to optimize the the patch with gradient descent. 

\noindent\textbf{Inverse Compositional (IC) Methods:}
One notable method in the tracking literature is Deep-LK~\cite{Wang17icra}, which also employed an \textit{iterative updating} approach using cascaded linear regression with a deep feature encoder. Their method is further extended by~\cite{Lv19cvpr_inverse_compositional} to enable 3D tracking in a $\text{SE}(3)$ manifold, whom also introduced learn-able optimization primitives to ease up the assumptions in the original IC formulation. By exploiting the IC method, these approaches can achieve both good accuracy and fast computation speed. However, such a non-probabilistic formulation also denied easy injection of domain specific knowledge into the optimization pipeline.

\noindent\textbf{Energy-based Models:}
Tracking with our factor graph consists of finding poses and associated appearances that minimize the sum of factors. Our approach inherits from energy-based models which learn an energy function that is low around observed data and is high otherwise \cite{LeCun06atutorial}. Although not explicitly energy-based, previous tracking methods have utilized Siamese correlation networks effectively to compute a response map over the tracked target's new position \cite{bertinetto2016fullyconvolutional,Wang19cvpr_siammask,tao2016siamese,zhang2020ocean}. This output response map does not require the calculation of a normalization constant like other probabilistic models and can be used to directly infer a tracked target's position to exploit the same advantage that energy-based models do. However, this generated response map is based on the forward pass of a deep learning model and does not easily allow for the addition of sources of information other than appearance. Our approach leverages the backward pass to optimize for an object's trajectory w.r.t. the energy at an observed pose.

%% file: 3-approach.tex
\section{Approach}

\subsection{Energy-based Tracking}
\begin{figure}[t]
\begin{center}
   \includegraphics[width=0.9\linewidth]{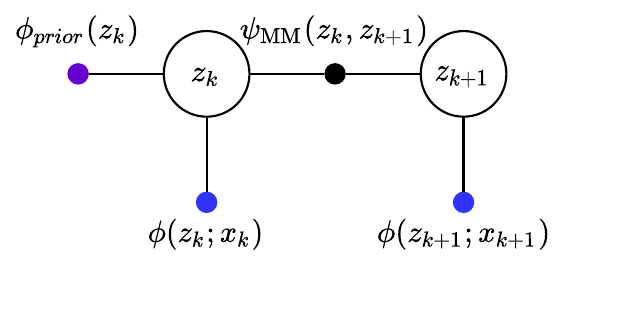}
\end{center}
\vspace{-2em}
\caption{Trajectory optimization factor graph. Purple: Prior Factor; blue: tracking factor; black: between factor.\vspace{-1em}}
\label{fig:fg}
\end{figure}
We model the state $\state_k$ as consisting of a pose $\pose_k \in \poses$ and appearance $\app_k \in \apps$, although our approach generalizes to adding more state, e.g., velocities to accommodate more complex motion models.
We assume that $\poses$ is an n-dimensional Lie group, e.g., $SE(2)$ or $SE(3)$, so that any pose $\pose$ can be written using the exponential map as $g=\exp(\xi)$ with $\xi\in \RR^n$ an element of the Lie algebra.

We formulate the problem as a factor graph (Fig. \ref{fig:fg}), with the hidden states $\state_k$ as the unknown variables, and with factors expressing the negative log-likelihood over the combination of variables they are connected to. 
We then recover the optimal incremental trajectory $\traj^*$ as the minimizer of
\begin{align}
\label{eq:factor_energy}
\traj^* = \arg \min_{\traj} \sum_i \factor_i(\traj_i),
\end{align}
where $\traj_i$ is the set of variables connected to factor $\factor_i$.

In tracking the factor graph is a simple Markov chain, with binary factors $\binary(\state_k,\state_{k+1})$ for the motion model, and unary factors $\unary(\state_k; \meas_k)$  modeling the likelihood of the state $\state_k$ given frame $\meas_k$. 
A simple way to model the motion is as a simple Brownian motion, using
\begin{align}
\binary(\pose_k,\pose_{k+1}) &= \| \log(\pose_k^{-1}\pose_{k+1}) \|^2_Q
\end{align}
where $\log$ is the inverse of the exponential map, $Q$ is an $n \times n$ covariance matrix, and $\| e \|^2_Q$ denotes the squared Mahalanobis error with covariance $Q$.

The time evolution of the appearance variables $\app_k$
can be modeled via a generic binary factor,
\begin{align}
\binary(\app_k,\app_{k+1}) = - \log p(\app_{k+1}|\app_k),
\end{align}
e.g., encoding Markov transition probabilities in the discrete case, or a continuous random walk in the case of continuous variables.

Note here that there is no limit to how many additional data sources could be added to the factor graph. Data sources could include  GPS, physical markers \cite{Mathis2018DeepLabCutMP} or Lidar. Additionally there is no  limit to how elaborate the motion could be. Modelling velocity, acceleration and jerk can be done in this tracking framework.

\subsection{The Likelihood Model}

\begin{figure}[t]
\begin{center}
   \includegraphics[width=0.9\linewidth]{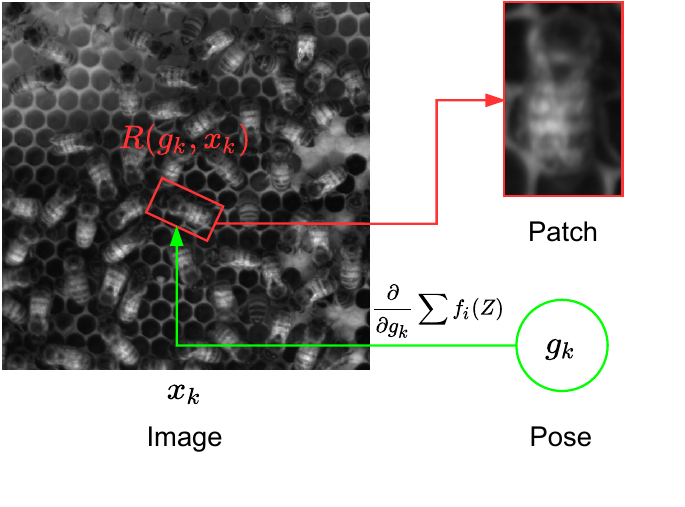}
\end{center}
\vspace{-2em}
\caption{Central to our approach is a differentiable warp $\region$ that extracts a region of interest from the image $\meas_k$ around pose $\pose_k$.}
\label{fig:warp}
\end{figure}

\noindent A key contribution is formulating the unary likelihood model $\unary(\state_k; \meas_k)$ as the likelihood ratio for a region $\region(\pose_k,\meas_k)$, obtained through a differentiable warp $\region$. In particular, $\region$ extracts a patch from a region of interest (oriented bounding box) from the image $\meas_k$ around pose $\pose_k$:
\begin{align}
    \label{eq:likelihood}
    \unary(\state_k; \meas_k) &= - \log \frac{p_F(\encoder(\region(\pose_k,\meas_k))|\app_k)}{p_B(\encoder(\region(\pose_k,\meas_k))))} \\
     &= \unary_F(\pose_k, \app_k; \meas_k) - \unary_B(\pose_k; \meas_k).
    \label{eq:split}
\end{align}
Above $p_F$ is a conventional probabilistic foreground model and $p_B$ is a conventional probabilistic background model, but the mapping $\encoder$ is a data-driven feature encoding from the image region to a d-dimensional feature vector $\feat \in \RR^d$. 
The specific form of \eqref{eq:likelihood} is informed by~\cite{Schonborn15cviu_bg_modeling.pdf}, in which the authors emphasized the importance of explicitly modeling the background in generative modeling.

In Equation \eqref{eq:split} we explicitly write this as a foreground factor $\unary_F$ and background factor $\unary_B$ to emphasize the factor graph structure. Because they are obtained by taking the negative log of a probability density, they can be interpreted in an energy minimization framework as illustrated graphically in Figure~\ref{fig:idea}.

The feature encoding $\encoder(.)$ itself can be learned, introducing deep trainable factors into a probabilistic trajectory optimization framework. The three encodings we evaluated are detailed in Section~\ref{sec:encodings}.

\subsection{Generative Models over the Feature Space}

For the foreground model $p_F$ and background model $p_B$ we have used both a full covariance Gaussian model and a na\"{i}ve Bayes model (a Gaussian model with a diagonal covariance matrix). In either case, we have
\begin{align}
\unary_{\{F, B\}}(\pose; \meas) &\define -\log p_{\{F, B\}}(\feat) \\
 &= -\log \Normal(\feat; \mu_{\{F, B\}}, \Sigma_{\{F, B\}}),
\end{align}
with $\feat=\encoder(\region(\pose,\meas))$ and
where $\mu_{\{F, B\}}$ and $\Sigma_{\{F, B\}}$ are the mean and covariance matrix for the foreground and background respectively.

Both foreground and background model are trained in a supervised manner on the extracted feature vectors $\feat=\encoder(\region(\pose,\meas))$, but for the background model we use arbitrary regions $\region(\pose^B_{ij},\meas_i)$ at poses $\pose^B_{ij}$ in the $i^{th}$  image that have minimal overlap with the foreground regions $\region(\pose^F_{ij},\meas_i)$.

\begin{algorithm}
\caption{Tracking by Optimization\label{algorithm:tracking}}
\begin{algorithmic}[1]
\State $K \gets $ number of video frames
    \For{k in 1...K}
    \State let $x_k =$ $k$-th image 
    \For{i in 1...N}
    \State sample pose $g_i$ from motion model $\psi(g_k, g_{k+1})$
    \State extract image patch $P_i=R(g_i,x_k)$
    \State compute error $E$ from image patch 
    \EndFor
    \State select best sample $g_b = \arg \min_{g_i}(E(g_i))$
    \State final pose $g_f = GradientDescent(g)$
    \EndFor
\end{algorithmic}
\end{algorithm}

\subsection{Optimization}
Computationally it is infeasible to compute $f_i(Z_i)$ for all possible $z_{k+1}$ states. Therefore, $f_i(Z_i)$ can only be computed for a subset of these possible states. Other tracking papers \cite{Wang19cvpr_siammask,siamRPN} use a region proposal network to constrain the possible target locations. Our method instead uses a motion model $\binary(\pose_k,\pose_{k+1})$ to provide initial guesses and then optimises the best guess. As shown in Algorithm~\ref{algorithm:tracking}, $N$ poses are sampled from the motion model and their error values $E(g)$ are computed. 
\begin{align}
    \label{eq:error}
    E(g) = \binary(\pose_k,\pose_{k+1}) + \unary(\state_{k+1}; \meas_{k+1})
\end{align}

As show in equation \eqref{eq:error}, the error value at a given state $z_{k+1}$  is a sum of the binary motion factor and the unary tracking factor. 

Then the pose with the lowest error $E(g)$ is further optimized with gradient descent. This part of the algorithm is one of the main contributions of this paper.
The derivative of the error $E$ is taken with respect to the Lie group parameters ($x,y,\theta$ in case of $\mathrm{SE}(2)$) and the pose is moved with respect to those parameters in the direction of steepest descent. This is performed until the pose converges to a local minimum. 

%% file: 4-encodings.tex
\section{Feature Encodings}\label{sec:encodings}

In the results section below we experiment with regularized auto-encoders (RAE, see~\cite{Ghosh20iclr_RAE}), probabilistic principal component analysis (PPCA, see~\cite{Tipping99ppca}), random projections, and Big Transfer~\cite{kolesnikov2019_bit}. We detail each of these below.

\subsection{Regularized Auto-encoders}

\newcommand{\cL}[1]{\mathcal{L}_{\mathbf{#1}}}
Regularized Auto-encoders (RAEs) are a recent development in obtaining a principled generative density over a domain, see~\cite{Ghosh20iclr_RAE} for more details. In this approach, we first train the auto-encoder backbone with the RAE loss
\begin{equation}
    \cL{RAE} = \cL{REC} + \cL{C}^{RAE} + \lambda\cL{REG},
\end{equation}
where $\cL{REC}$ is the reconstruction loss $\|\meas - D(\encoder(\meas))\|_2^2$, $\cL{Z}^{RAE}$ the regularizer for the code space ($\frac 1 2 \|c\|^2$), and $\cL{REG}$ is the explicit regularizer for the decoder parameter space ($\| \theta_D \|_2^2$ in our case).

We then obtain the feature encoding $\encoder(.)$ as the encoder of the RAE. In this case, $d$ is the dimension of the auto-encoder bottleneck. Finally, we fit the generative foreground model $p_F$, using either a single Gaussian or a mixture of Gaussians, i.e., we adopt the idea first proposed in~\cite{Ghosh20iclr_RAE}, to obtain a regularized auto-encoder (RAE). However, in order to use the RAE in the context of trajectory optimization, we also fit a background model, by using the same encoder $\encoder(.)$ and imputing a different density $p_B$.

\subsection{Probabilistic Principal Component Analysis}
\newcommand{\WW}{W}
\newcommand{\II}{I}
\newcommand{\CC}{C}
\newcommand{\RRR}{R}
\newcommand{\UU}{U}
\newcommand{\MM}{M}
\newcommand{\LLambda}{\Lambda}

A well known classical generative modeling technique is the Probabilistic Principal Component Analysis (PPCA) model by Tipping and Bishop~\cite{Tipping99ppca}. In PPCA, we assume that the data follows a linear factor analysis distribution,
\begin{equation}
    \meas = \WW c + \mu + \epsilon,
    \label{eq:factor-analysis}
\end{equation}
where $\meas$ is the observation, $c$ a latent code vector, and $\WW$ is a $d\times q$ matrix describing how the observation and latent variables relate. 
The mean $\mu$ allows the model to have non-zero mean
and $\epsilon \sim \mathcal{N}(0, \sigma^2\II)$ is zero-mean Gaussian noise with covariance $\sigma^2\II$.

The maximum likelihood estimates (MLE) of $\WW$ can be obtained in closed form using SVD \cite{Tipping99ppca}:
\begin{align}
   \WW_{ML} = \UU_q (\LLambda_q - \sigma^2 \II)^{1/2} \RRR,
\end{align}
and the MLE of $\sigma$, $\sigma^2_{ML} = \frac 1 {d-q} \sum_{j=q+1}^{d} \lambda_j$, where the $q$ columns of $\UU_q$ are principal eigenvectors of $S$ and their corresponding eigenvalues in $\LLambda_q$. Above $\RRR$ is an arbitrary $q\times q$ rotation matrix, and $\lambda_{q+1},\dots, \lambda_{d}$ are the eigenvalues ``lost'' in $W_{ML}$. Conversely, we have a density of $c$ over $\meas$:
\begin{equation}
   P(c | \meas) \sim \mathcal N (\MM^{-1}\WW^T (\meas - \mu), \sigma^2 \MM^{-1})
\end{equation}
where $M = {\WW}^T{\WW} + \sigma^2 \II$, which we can use to obtain an estimate for $c$ given a measured $\meas$.

\subsection{Random Projection Features}
Since random projections preserve data statistics with a high probability~\cite{Vempala04book_rp}, we can also use a random projection as a feature encoder.
In particular, let $A$ be an $n\times k$ random matrix whose entries are independently sampled from $\mathcal{N}(0, 1)$. For any arbitrary data vector $\meas \in \mathbb{R}^n$ the random projection $C(\meas)$ defined as
\begin{equation}
   c \define C(\meas) = \frac 1 {\sqrt k}A^T\meas
\end{equation}
preserves distance in the original feature space. 

Indeed, by Lemma 1.3 in~\cite{Vempala04book_rp} we have
\begin{flalign}
   \mathbf E (\| c\|^2) = \|\meas\|^2 \\
   P(\left| \| c \|^2 - \| \meas \|^2 \right| \geq \eta \|\meas \|^2) < 2 e^{-(\eta^2 - \eta^3)k/4}
\end{flalign}
for any $\eta > 0$. Thus when $k$ is sufficiently large, we can expect that the random project will preserve the $L_2$ distance of vectors with a high probability. Consequently, we can use the random projection above as an encoder, under the assumption that densities fitted to this lower-dimensional feature space will model densities in the original image space. Because this encoding $C$ is \textit{not} learned, unlike RAE and PPCA, it provides a good baseline model.

\subsection{Big Transfer}
In addition to feature encodings, we also experiment with image classification to directly infer the probability of a given image patch being a foreground $p_F$ or background $p_B$ image. We evaluate Big Transfer~\cite{kolesnikov2019_bit} in our results which is outlined below.

Big Transfer (BiT) is a training paradigm that leverages a pre-trained and modified ResNet-v2~\cite{He016_identitymappings} architecture along with data-specific training parameters that have achieved state-of-the-art results on a series of downstream tasks~\cite{kolesnikov2019_bit}. In particular, all Batch Normalization~\cite{ioffe2015batch} layers in the ResNet model are replaced with Group Normalization~\cite{Wu2018_groupnorm} layers and all convolutional layers implement Weight Standardization~\cite{Qiao2019_weightstandardization}. The BiT-M model, which we experiment with, is then trained on the ImageNet-21k~\cite{Deng2009_imagenet} dataset and is provided for training on downstream classification tasks.

To obtain our foreground and background classifier, we fine-tune the BiT-M model using the BiT-Hyperrule recommended learning rate schedule, resolution, and MixUp regularization~\cite{Zhang2017_mixup} on the downstream foreground and background bee hive image patches. This ResNet-v2 model with a softmax classification head then directly infers the probability $p$ of new image patches in our dataset being foreground $p_F$ or background $p_B$ images.

%% file: 5-results.tex
\section{Results}

\begin{figure*}
\begin{center}
\includegraphics[width=1.0\linewidth]{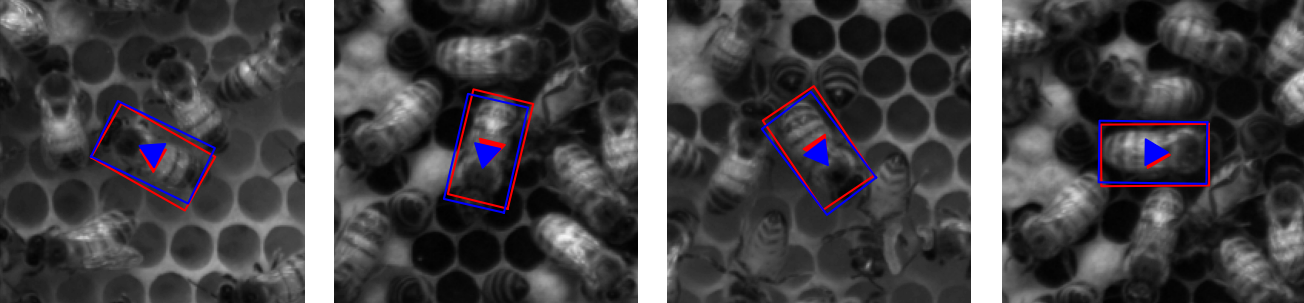}
\end{center}
\caption{Sample beehive images from OIST dataset from four of the twenty distinct tracks. Shown on each image are the bounding boxes oriented by the 2D pose of the bee, where the {\color{blue} blue} bounding box is the ground truth and the {\color{red} red} bounding box is the corresponding prediction from our algorithm. \vspace{-1em}}
\label{fig:short}
\end{figure*}
We evaluated our method on a challenging social insect dataset consisting of video of hundreds of bees moving over a bee hive grating, made public by OIST~\cite{Bozek20biorxiv_bee_tracking}.  In this case the space of poses $\poses$ is the Lie group $SE(2)$ of 2D Euclidean rigid transforms, and the differentiable warp $R$ extracts an oriented bounding box centered on the pose $\pose_k\define(x_k,y_k\theta_k)$. We iteratively optimize the factor graph at each time $\meas_k$ to recover the state $z_k$ on a set of eighty frames for each of twenty labeled bee tracks. 

\subsection{Dataset and Evaluation Metrics}
The OIST dataset contains a set of  bee annotations that have been hand labelled over hundreds of frames from a video of the bee hive shown in Figure \ref{fig:short}. Each video frame has a resolution of $2560 \times 2560$ with $70 \times 40$ labeled bounding boxes. For training, the feature encoding $\encoder(.)$ is trained through reconstructing the region $\region$ enclosed within an oriented bounding box for a given pose $\pose_k$ for the RAE, RP, and PPCA models. For the BiT approach, the model is trained to directly infer the probability that a given image patch is foreground $p_F$ or background $p_B$. These models are trained on 3,000 foreground and 3,000 background bee images present within the first 100 frames of the dataset. The image patches are then evaluated on a hold-out test set of 20 tracking sequences each with eighty frames.

We report tracking performance as the Expected Average Overlap (EAO), which is a function of the accuracy and robustness of a tracker over all tracking sequences in a dataset~\cite{Cehovin_Zajc2016a}. We also show the average accuracy and robustness over all sequences as individual metrics, where accuracy is defined as the average overlap between the predicted and ground truth oriented bounding boxes before failure, and robustness is the proportion of frames where the tracker has succeeded in tracking the correct object. These metrics are calculated over eighty frames used for tracking in the hold-out test dataset.

\subsection{Implementation Details}

We first train a feature encoding $\encoder(.)$ on the oriented bounding boxes centered on each pose $\pose_k$ for all labelled insects in the first one hundred frames. For the RP, PPCA, and RAE models, we then fit a foreground model $p_F$ and background model $p_B$ to create the foreground factor $\unary_F$ and background factor $\unary_B$ trained on these poses. We found that modeling both $p_F$ and $p_B$ as multivariate Gaussian generative models obtained the best results. With BiT we use the softmax output to directly infer the probability of an image patch being foreground $p_F$ or background $p_B$. To infer the MAP trajectory $\traj^*$ on a given sequence, we iteratively add measurements for each frame at time $t_k$ to a factor graph as shown in Figure \ref{fig:fg} and optimize to recover the hidden state $\state_k$ of the object across the track. This inference is conducted across eighty frames for all twenty tracks to obtain the MAP trajectory $\traj^*$ across all sequences in the test dataset. 

We evaluated Random Projection (RP), Probabilistic Principal Component Analysis (PPCA), Regularized Auto-encoder (RAE), and Big Transfer (BiT)  models to determine their performance in imputing a useful set of densities $p_F$ and $p_B$ for the factor graph to optimize on. The RAE model is constructed with a single convolution and max pool layer layer followed by two fully connected layers which are then up-sampled in an equivalent inverse decoder step. The auto-encoder is penalized using reconstruction loss on the foreground image patches and trained using the Adam~\cite{kingma2014adam} optimizer with a learning rate of $0.001$. Each auto-encoder is trained outside of the factor graph optimization loop and the trained weights are loaded in for image patch encoding when the deep factors are initialized. The BiT training process follows the procedures outlined in BiT-Hyperrule~\cite{kolesnikov2019_bit} for the BiT-M model pre-trained on ImageNet-21k.

The entire trajectory optimization pipeline is implemented with Swift for TensorFlow (S4TF, see~\cite{saeta2021swift}), allowing for a simple implementation of auto-differentiation across the entire factor graph optimization process.  

\subsection{Deep Features Outperform Linear Features}
\begin{figure*}
\begin{center}
   \includegraphics[width=1.0\linewidth]{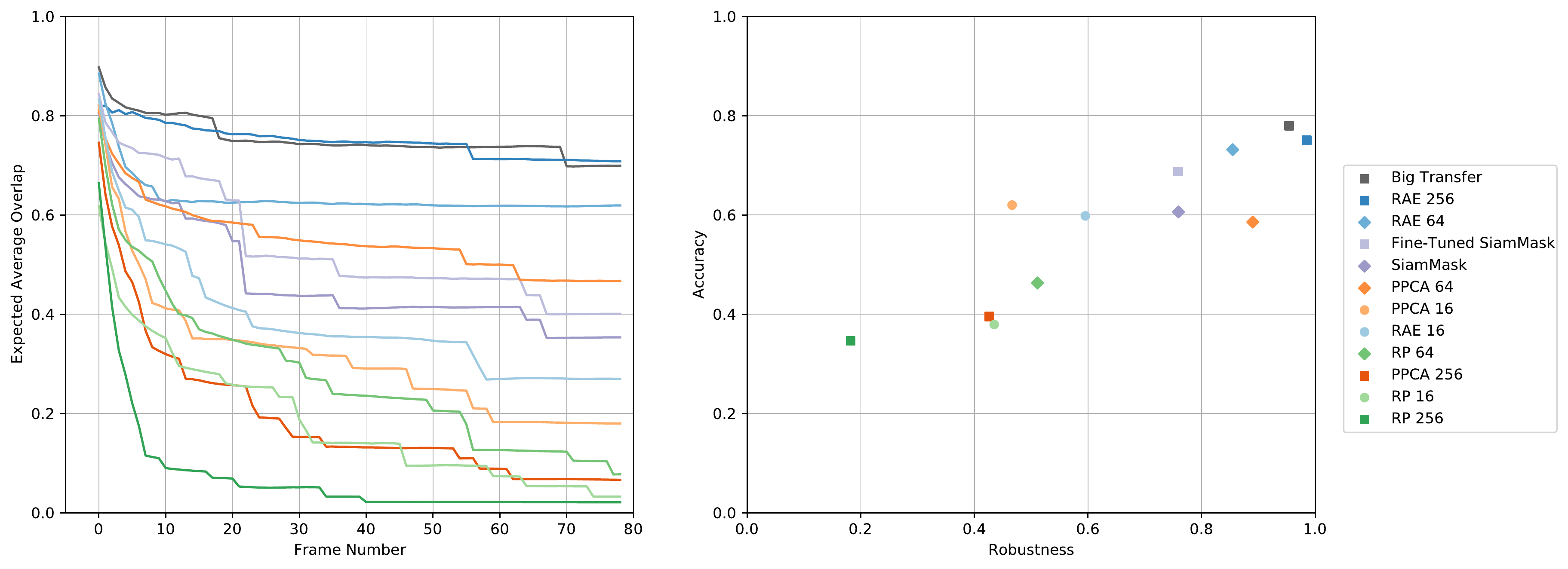}\vspace{-1em}
\end{center}
\caption{Performance of the BiT~\cite{kolesnikov2019_bit}, RAE, PPCA, RP Factor Graphs alongside the state-of-the-art model SiamMask~\cite{Wang19cvpr_siammask} on the test dataset. The EAO curve for each model (left) and the AR-plot (right) are calculated over all twenty object tracking sequences for eighty frames. Each factor graph configuration is listed by its appearance encoder and the dimensionality of the encoded output $d \in \{16, 64, 256\}$.
\label{fig:tracker_comparison}\vspace{-1em}}
\end{figure*}
In the first set of results, we show that using deep factors from BiT and RAE models outperform both a linear PPCA appearance model~\cite{Khan04cvpr} and a linear feature set obtained using random projections (RP, see ~\cite{Vempala04book_rp}). We also show that the RAE models with a higher number of output features achieve higher EAO than their counterparts with fewer dimensions. Furthermore, we find that the BiT approach yields higher tracking accuracy whereas the RAE 256 model achieves greater robustness.

To determine the efficacy of deep factors, we ran a series of experiments to find the EAO performance of RP, PPCA, and RAE models each with bottleneck sizes $d \in \{16, 64, 256\}$ alongside the BiT approach and found that the BiT and RAE 256 approaches outperformed every other factor graph configuration as seen in Figure \ref{fig:tracker_comparison}. The RAE model with bottleneck size $d=256$ outperformed the linear feature encodings with $0.75$ accuracy and $0.98$ robustness while the BiT approach achieved the highest accuracy $0.78$ with comparable robustness $0.95$ to the RAE 256 approach. The top end EAO scores for each model increase from 0.27 EAO for RP to 0.55 for PPCA to 0.75 for RAE. Across each output size, the RAE and BiT models achieved significantly better results, demonstrating the effectiveness of deep factors.

Another important finding is that increasing the bottleneck size $d$ of the RAE model significantly increases EAO performance. As the bottleneck dimensionality $d$ of the RAE increases from 16 to 256, performance significantly improves from 0.39 EAO score to 0.75. We also found that for bottleneck size above the 256 threshold $d > 256$, EAO performance stayed relatively constant. This demonstrates that increasing the complexity of the RAE model yields significant gains in its ability to find an effective feature encoding $\encoder(.)$ that efficiently condenses high dimensional inputs, resulting in increased EAO for these larger bottleneck dimensions $d \geq 256$.

\subsection{Deep Factors Outperform Deep Trackers}

The RAE 256 and BiT deep factor graph approaches obtained higher EAO, accuracy, and robustness than SiamMask, a state-of-the-art short-term object tracker, across all test sequences. SiamMask~\cite{Wang19cvpr_siammask} is a representative of a family of Siamese trackers that can produce oriented bounding boxes by estimating a minimal area rectangle around a segmentation mask~\cite{ma2020rpt,zhang2020accurate,yan2020alpharefine}. 

We evaluated the performance of SiamMask using a model trained solely on VOT2019 as well as the same model fine-tuned on the OIST dataset. SiamMask comes with models pre-trained on either VOT2019~\cite{Kristan2019a} or DAVIS2016~\cite{Perazzi2016} datasets and the VOT2019 model is slightly better performing on the OIST testing dataset. SiamMask takes a single-bounding box at initialization, producing oriented bounding boxes and object segmentation masks. SiamMask is made up of an offline-trained fully-convolutional Siamese network~\cite{bertinetto2016fullyconvolutional}. At inference, a larger $255\times255\times3$ search image from the current frame and a smaller $127\times127\times3$ target image from the initial frame are both passed into the same CNN. This yields two dense feature maps which are depth-wise cross-correlated. The result is a response map which will indicate where in the search image the object is most likely located.

\begin{table}
\begin{center}
\begin{tabular*}{1.0\linewidth}{@{\extracolsep{\fill}}ccccc}
\hline
& $d$ & Accuracy & Robustness & EAO\\ 
\hline\hline 
{BiT} & \textbf{BiT-M} & \textbf{0.78} & \textbf{0.95} & \textbf{0.75}\\
\hline
{RAE} & \textbf{256} & \textbf{0.75} & \textbf{0.98} & \textbf{0.75}\\
& 64 & 0.73 & 0.85 & 0.63\\
& 16 & 0.60 & 0.60 & 0.39\\
\hline
{PPCA} & 256 & 0.40 & 0.43 & 0.19\\
& 64 & 0.59 & 0.89 & 0.55\\
& 16 & 0.62 & 0.47 & 0.31 \\
\hline
{RP} & 256 & 0.35 & 0.18 & 0.07\\
& 64 & 0.46 & 0.51 & 0.27\\
& 16 & 0.38 & 0.43 & 0.18 \\
\hline
\end{tabular*}
\end{center}
\caption{The accuracy and robustness of the feature encoding $\encoder(.)$ configurations from our experimental results with encoding output dimensionality $d \in \{16, 64, 256\}$. For consistency, all models leveraged multivariate Gaussian models for both the foreground $p_F$ and background $p_B$ generative densities. The BiT model implements the default BiT-M ResNet-v2 model architecture~\cite{kolesnikov2019_bit}. }
\label{fig:deep_sota_table}
\end{table}
\begin{figure}
\begin{center}
\includegraphics[width=1.0\linewidth]{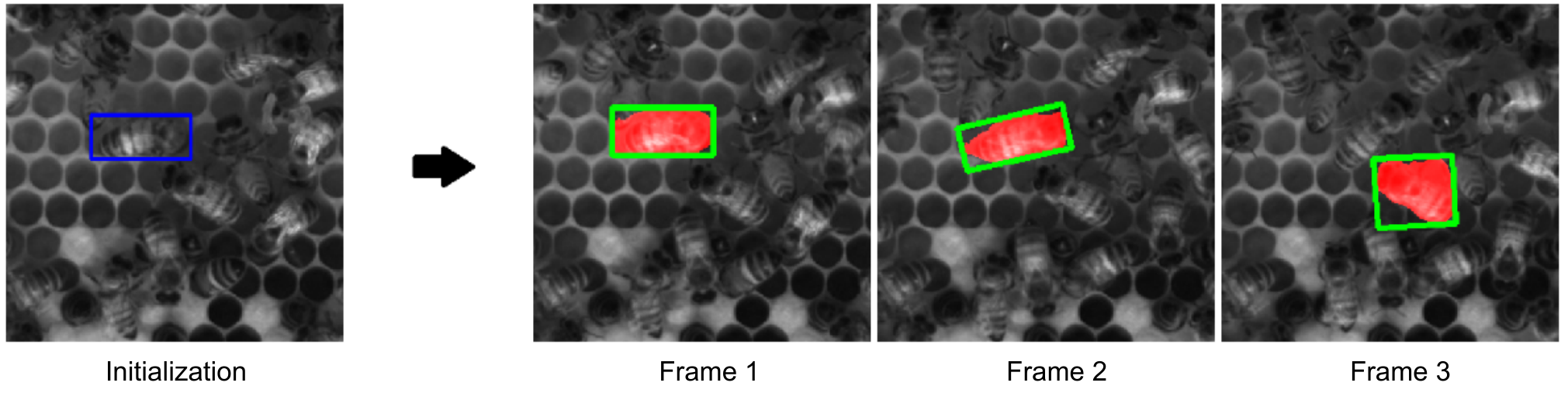}\vspace{-1em}
\end{center}
\caption{SiamMask fails to keep track of the correct bee and jumps to an adjacent bee. Bounding box initialization (left). Output segmentation masks and oriented bounding boxes of the next three frames (right).
\label{fig:siammask_failure}\vspace{-1em}}
\end{figure}
In Figure \ref{fig:tracker_comparison} we show that the the RAE 256 and BiT trackers outperform both SiamMask and SiamMask fine-tuned on the OIST dataset.  The RAE tracker achieves higher accuracy and robustness than the better performing fine-tuned SiamMask model, increasing accuracy from 0.69 to 0.75 and robustness from 0.76 to 0.98. Similarly, the BiT approach improves accuracy from 0.69 to 0.78 and robustness from 0.76 to 0.95.  Because of the nature of this dataset there are multiple similar looking objects in each frame. SiamMask's Siamese network will yield a high response value to objects that look very similar to the target object which appear close to the target location in the frame. This can cause the tracker's prediction to jump to nearby bees due to the very crowded environment of a beehive, reducing the robustness of that sequence.

The fine-tuned SiamMask model outperformed the model trained solely on the VOT dataset, increasing accuracy from 0.61 to 0.68.  The robustness, however, remained constant at 0.76 despite the additional domain-specific training. The training data used for fine-tuning is comprised of the first one hundred frames of the OIST dataset as well as twenty tracks containing the oriented bounding box coordinates for twenty different bees. We converted the training data to match the input required for SiamMask training. Oriented bounding boxes were converted to segmentation masks by constructing oriented ellipses with the same center, angle, width, and height of the oriented bounding boxes.

\subsection{Differentiable Oriented Bounding Boxes}

\begin{figure}[t]
\begin{center}
   \includegraphics[width=1.0\linewidth]{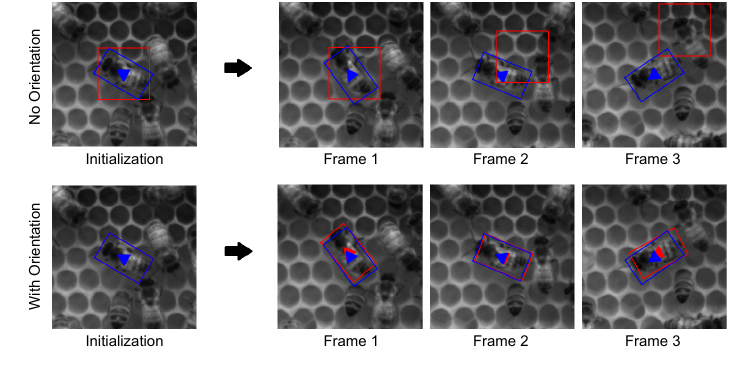}\vspace{-1em}
\end{center}
\caption{RAE 256 factor graph trained with a differentiable bounding box (top) and an differentiable oriented bounding box (bottom) where the prediction is shown in {\color{red} red} and the ground truth in {\color{blue} blue}. Explicitly modeling the orientation through an oriented bounding box allows for the factor graph to correctly map the trajectory of the rotating bee, whereas by modeling just translation we see that as the most likely bounding box chosen was incorrect as the bee rotates.\label{fig:diff_obbs}\vspace{-1em}}
\end{figure}
Explicitly modeling orientation through an oriented bounding box performs better than modeling the orientation as part of the appearance variables $\app_k$. Figure \ref{fig:diff_obbs} shows a situation where a RAE 256 model trained with an oriented differentiable bounding box outperforms a RAE 256 model with a bounding box that does not model orientation. We see that as the bee with appearance $a_k$ rotates near another bee with similar appearance $\hat{a}_k$, the factor graph with the oriented bounding box is able to correctly reconstruct the trajectory $\traj^*$ while the bounding box without orientation fails by switching to the bee with appearance $\hat{a}_k$. The explicit orientation information at time $t_k$ allows the factor graph to optimize with better constraints on the possible pose $g_{k + 1}$ of the bee in the next time step $t_{k + 1}$,

In general, the tracker that is able to optimize the explicit orientation of the bounding box attains higher robustness and accuracy across all sequences because of the additional orientation information. Using the RAE 256 model with oriented bounding boxes improves accuracy from 0.39 to 0.75 and robustness from 0.33 to 0.98 averaged across all tracks. Furthermore, when orientation is not captured, a $70 \times 70$ square bounding box must be used to ensure that the $40 \times 70$ bee will be encapsulated by the bounding box. This reduces the specificity of the of the object tracker which is also less desirable as it does not capture the exact location of the object. It also increases the complexity of the feature encoding $\encoder(.)$ by increasing the size of the input image patch. Overall, explicitly modeling the orientation through a differentiable oriented bounding box provides more consistent and precise results than attempting to capture orientation information through the appearance $\app_k$.

\subsection{Choice of Generative Model is Important}

\begin{figure}[t]
\begin{center}
   \includegraphics[width=0.85\linewidth]{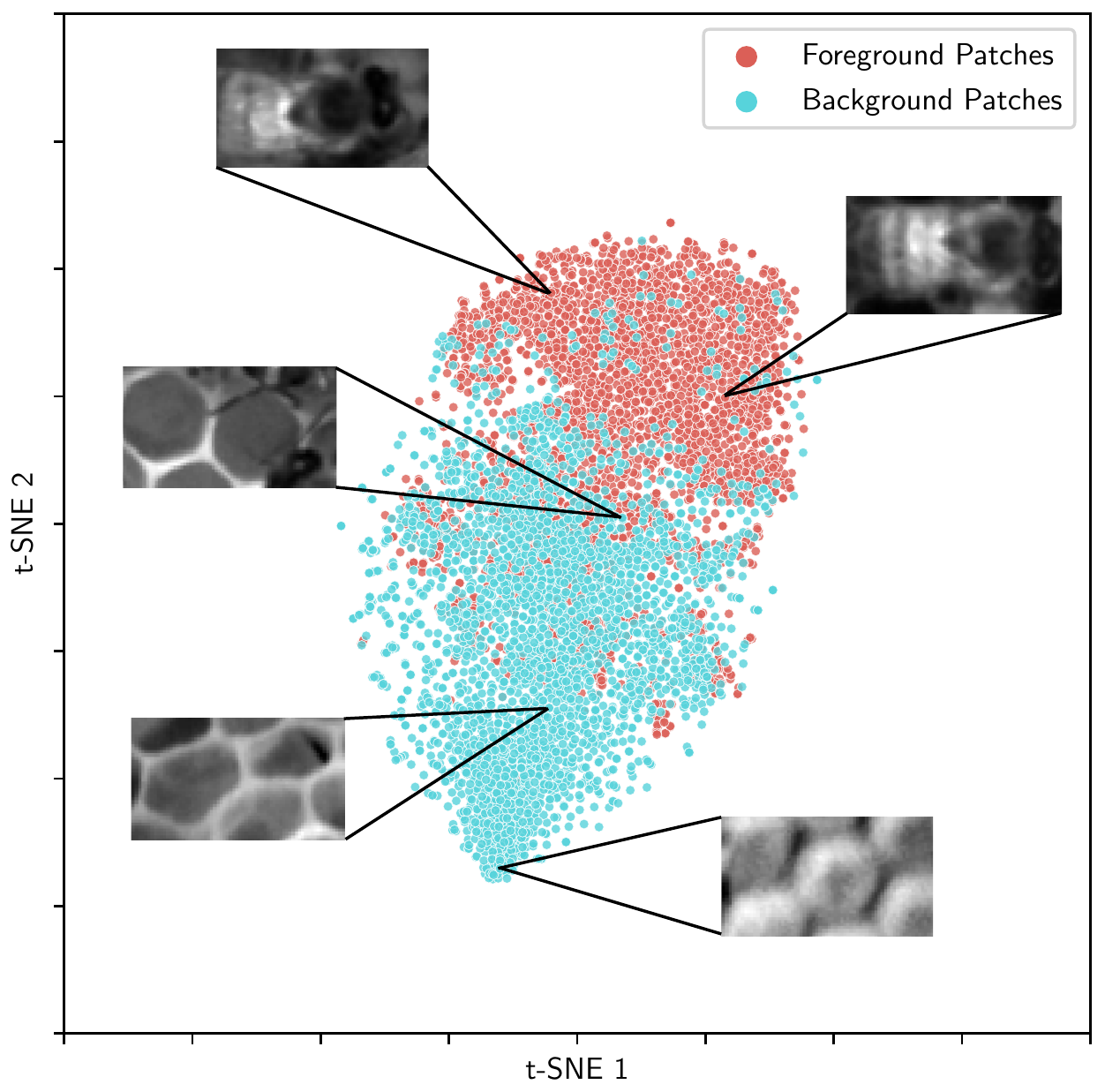}
\end{center}
\caption{The two component t-SNE~\cite{Vandermaaten08jmlr} visualization of the encoded image patches using the RAE 256 feature encoding $\encoder(.)$ for six thousand foreground and background images. Encoded features $\encoder(\region_F)$ and $\encoder(\region_B)$ have a strong grouping in the two t-SNE dimensions, indicating that a mixture of Gaussian models would not represent the data well in the higher encoded dimension $d=256$. Furthermore, we validate the need to represent the encoded foreground and background regions with two separate generative densities $p_F$ and $p_B$ as they separate well in the two component t-SNE output. The majority of the overlap between the two encoded regions are those background regions $\region_B$ which do contain some part of a bee in the periphery of the image.\label{fig:tsne}\vspace{-1em}}
\end{figure}
In this experiment, we establish that using a set of multivariate Gaussian generative densities for the foreground model $p_F$ and background model $p_B$ achieves higher EAO than using a multivariate Gaussian foreground model $p_F$ and a na\"{i}ve Bayes background model $p_B$ with a RAE 256 feature encoding $\encoder(.)$. In this case, we consider a multivariate Gaussian model for the foreground density $p_F$ and both a multivariate Gaussian model and a na\"{i}ve Bayes model for the the background density $p_B$. We show that using a full covariance Gaussian for the background model increases accuracy by 0.31 and robustness by 0.72 as opposed to the na\"{i}ve Bayes model when using the RAE 256 feature encoding $\encoder(.)$. Based on these results, it is clear that the full $d \times d$ covariance matrix in the multivariate Gaussian model is necessary to capture the relationship between features of the RAE 256 encoded background regions $\encoder(\region_B)$. 

In Figure \ref{fig:tsne}, we also illustrate that the foreground and background models $p_F$ and $p_B$ are not represented well by a mixture of Gaussian models in the two component plane output by t-SNE~\cite{Vandermaaten08jmlr}, indicating that a Gaussian mixture also would not represent the data well in the higher RAE 256 encoded dimension $d=256$. We applied the RAE 256 feature encoding $\encoder(.)$ on six thousand foreground and background regions $\region$ and reduced the encoding dimensionality $d=256$ to two $d=2$ using t-SNE with perplexity 35. By analyzing the output from these two components, it is clear that within the foreground regions $\region_F$ and background regions $\region_B$, there are no significantly separated groups that would imply the need for a mixture of Gaussian models. Alongside this finding, we also validate the necessity for splitting the encoded foreground regions $\encoder(\region_F)$ and encoded background regions $\encoder(\region_B)$ into separate generative densities $p_F$ and $p_B$, as it is clear they do not exhibit the same encoded output from the RAE 256 model.

%% file: 6-conclusions.tex
\section{Conclusions}

Using a deep feature encoding in conjunction with generative models over the features in a factor-graph based, probabilistic trajectory optimization framework yields excellent results in a challenging application.
Our method leverages the Lie group properties of the tracked target's pose by using a differentiable warp, which is optimized in a structured probabilistic optimization framework.

This can form the basis for more complicated applications. Above we experimented with $\mathrm{SE}(2)$, and it would be of interest to see the results for more general poses, e.g., 2D similarity transforms and full 3D poses, similar to~\cite{Lv19cvpr_inverse_compositional}.

While we obtain good results using the auto-encoder features, the RAE is trained for image reconstruction rather than performance on the trajectory optimization task. Training the auto-encoder with the final task, i.e., with the optimization in the loop, is the subject of future work.
Finally, while our current approach uses a supervised, off-line training method, but nothing prevents on-line adaptation of either features and the generative foreground model.